\documentclass{IOS-Book-Article}

\usepackage{mathptmx}
\usepackage{soul}\setuldepth{article}
\usepackage{graphicx}
\usepackage[table]{xcolor}

\def\hb{\hbox to 11.5 cm{}}

\begin{document}

\pagestyle{plain}

\begin{frontmatter}              

\title{Data-Centric Machine Learning in the Legal Domain}

\markboth{}{January 2022\hb}

\author[A]{\fnms{Hannes} \snm{Westermann}
\thanks{Corresponding Author: Hannes Westermann, E-mail: hannes.westermann@umontreal.ca}},
\author[B]{\fnms{Jarom\'{i}r} \snm{\v{S}avelka}},
\author[C]{\fnms{Vern R.} \snm{Walker}},
\author[D]{\fnms{Kevin D.} \snm{Ashley}} and
\author[A]{\fnms{Karim} \snm{Benyekhlef}}

\runningauthor{H. Westermann et al.}
\address[A]{Cyberjustice Laboratory, Facult\'e de droit, Universit\'e de Montréal}
\address[B]{School of Computer Science, Carnegie Mellon University}
\address[C]{LLT Lab, Maurice A. Deane School of Law, Hofstra University}
\address[D]{School of Computing and Information, University of Pittsburgh}

\begin{abstract}
Machine learning research typically starts with a fixed data set created early in the process. The focus of the experiments is finding a model and training procedure that result in the best possible performance in terms of some selected evaluation metric. This paper explores how changes in a data set influence the measured performance of a model. Using three publicly available data sets from the legal domain, we investigate how changes to their size, the train/test splits, and the human labelling accuracy impact the performance of a trained deep learning classifier. We assess the overall performance (weighted average) as well as the per-class performance. The observed effects are surprisingly pronounced, especially when the per-class performance is considered. We investigate how ``semantic homogeneity'' of a class, i.e., the proximity of sentences in a semantic embedding space, influences the difficulty of its classification. The presented results have far reaching implications for efforts related to data collection and curation in the field of AI~\&~Law. The results also indicate that enhancements to a data set could be considered, alongside the advancement of the ML models, as an additional path for increasing classification performance on various tasks in AI \& Law. Finally, we discuss the need for an established methodology to assess the potential effects of data set properties.
\end{abstract}

\end{frontmatter}
\markboth{January 2022\hb}{January 2022\hb}

\section{Introduction}
This paper is an extended version of \cite{westermann2021}.
Two fundamental components of a machine learning (ML) experiment are data and a model. The ML community appears to prefer putting more effort into tweaking the models while spending less time on important data considerations \cite{nithya2021}. This means that researchers often invest considerable resources into developing novel models and approaches, achieving marginal improvements. At the same time, they pay much less attention to the properties of the data set (e.g., size, quality, train/test split), or to the effects these might have on the performance of the fitted ML models. Potentially, this underinvestigated area of research could lead to significant improvements in the performance of the models.

We investigate the effects of changing data set properties on the performance of a trained classifier in three different experiments: expanding the size of the data set, selecting particular splits for training and test sets, and decreasing the quality of annotation. This may be very important for legal data sets, which are often small because creating them tends to be costly and time-consuming. 
Our experiments suggest that analyzing how data set properties affect performance can be an important step in improving the results of trained classifiers, and in better understanding the obtained results. For example, our results show that collecting additional data is likely to improve performance for two out of the three studied data sets, which could inform the decision of a researcher to continue collecting data, and that focusing data collection on certain difficult classes might improve overall performance. We investigate the use of ``semantic homogeneity'', i.e., the proximity of sentences in a semantic embedding space, to identify such difficult classes.

\section{Related Work}


Recently, there has been an increased interest in the general ML community to explore how data set properties affect the performance and characteristics of trained classifiers. In \cite{nithya2021}, the researchers investigated data-cascades, where issues with the labelling of data affected downstream systems in high-stakes domains. In \cite{northcutt2020}, the authors estimated that the ten of the most commonly used ML data sets contain an average of 3.4\% errors in labelling. In \cite{sureshguttag2020}, the authors identified six ways that bias could enter into the ML pipeline and lead to unintended consequences. Several of these are related to the data generation process. In this paper, we investigate how such data issues affect the downstream performance of a trained ML system. 

Researchers in AI \& Law have explored manipulating data sets to assess effects on model performance as well. For example, \cite{zhong2019} experimented with iterative masking of the most predictive sentences. Other examples include ablating data about criminal charges or sentences \cite{tan2020}, or adding ODR dispute data to lawsuit data to improve judgment prediction \cite{zhou2020}.
In \cite{savelka2020}, the authors investigated how recasting multiple data sets into a single task and pooling them in different combinations affected the performance of the trained classifier. The authors of \cite{savelka2021} annotated judicial decisions from eight languages, jurisdictions and domains, in terms of their functional parts and investigated how training a classifier on a single or multiple data sets impacted a model's performance on other data sets. The classifier was able to learn and benefit from data in other languages, jurisdictions and domains. Here, we draw different distributions from a data set and alter the data to observe the effects on the classifier's performance.

In \cite{westermann2020}, the authors studied how pre-trained sentence encoders placed sentences of the same class in close proximity in a vector space. They found that many of the classes had a surprisingly high ``semantic homogeneity'', with surrounding sentences  in the vector space all having the same label. Here, we employ the same data sets to investigate how changing the quantity, training/test split, and labelling quality affects the model's per-label performance.

\section{Experimental Design}
We evaluated a deep learning classifier (section \ref{sec:model}) on three publicly available data sets annotated on a sentence level (Section \ref{sec:data}) under three experimental settings (Section \ref{sec:experiments}). We investigated how changes to the data sets affect the classifier's performance. The code to carry out the experiments is available on github.\footnote{\url{https://github.com/hwestermann/jurix2021-data_centric_machine_learning}}

\subsection{Data}
\label{sec:data}
In \cite{walker:2019} the researchers analyzed 50 fact-finding decisions of the U.S. Board of Veterans' Appeals \textbf{(BVA)} (2013 - 2017). All arbitrarily selected  cases dealt with claims by veterans for service-related post-traumatic stress disorder (PTSD). The researchers extracted 6,153 sentences addressing the factual issues related to the claim for PTSD or related psychiatric disorders. These were tagged with  rhetorical roles \cite{walker:2017} the sentences play in the decisions: Finding, Reasoning, Evidence, Legal Rule, Citation and Other.\footnote{Dataset available at \url{https://github.com/LLTLab/VetClaims-JSON}}

In \cite{savelka2019} the authors studied retrieving sentences from court opinions that elaborate on the meaning of vague statutory terms \textbf{(StatInt)}. Their data set contains sentences from case law that mentioned three terms from different U.S. Code provisions. They manually classified the sentences in terms of four categories of usefulness for explaining the corresponding statutory term. Our second data set contains the sentences mentioning `common business purpose' (149 high value, 88 certain value, 369 potential value, 274 no value). In \cite{savelka2019} the goal was to rank the sentences with respect to their usefulness; here, we classify them into the four value categories.\footnote{Data set available at \url{https://github.com/jsavelka/statutory_interpretation}}

The third data set also focuses on rhetorical roles of sentences. Bhattacharya et al.~\cite{bhattacharya2019} analyzed 50 opinions of the Supreme Court of India \textbf{(ISC)}. They sampled cases from five  domains in proportion to their frequencies (criminal, land and property, constitutional, labor and industrial, and intellectual property). The decisions were split into 9,380 sentences and  classified into one of  seven categories of rhetorical roles they play in a decision: Facts, Ruling (lower court), Argument, Ratio, Statute, Precedent, Ruling (present court). The data set was used as a sequence labelling data set in \cite{bhattacharya2019}. Here we  recast it as a sentence classification problem, without taking surrounding sentences into account.\footnote{Data set available at \url{github.com/Law-AI/semantic-segmentation}}

\begin{table}[]
    \centering
    \setlength{\tabcolsep}{9pt}
    \begin{tabular}{lr|lr|lr}
              \multicolumn{2}{c|}{\cellcolor{black!8}BVA} & \multicolumn{2}{c|}{\cellcolor{black!8}StatInt} & \multicolumn{2}{c}{\cellcolor{black!8}ISC} \\
              
              Label & freq. & Label & freq. & Label & freq.\\
              \hline
                Other &  477 & No Value & 274 & Facts & 2219 \\
                Finding &  490 & Potential & 369 & Ruling (lower) & 316 \\
                Evidence &  2420 & Certain & 88 & Argument & 845 \\
                Rule &  938 & High & 149 & Ratio & 3624 \\
                Citation &  1118 & &  & Statute & 646 \\
                Reasoning &  710 &  &  & Precedent & 1468 \\
                &   &  &  & Ruling (present) & 262 \\
                \hline
                Total & 6153  &  &  880 &  &  9380\\
	 
    \end{tabular}
    \caption{The distribution of labels for each dataset}
    \label{tab:per_label}
\end{table}

\subsection{Model}
\label{sec:model}
We embed  each sentence in the data set using a pre-trained transformer architecture \cite{vaswani:2017} and deep averaging network \cite{iyyer2015}, the Google Universal Sentence Encoder \cite{cer2018} (\textbf{GUSE}).\footnote{\url{https://tfhub.dev/google/universal-sentence-encoder/4}}. In \cite{westermann2020} this model  efficiently placed sentences belonging to the same class, as defined by some annotation schema, together in a high-dimensional vector space.

We input these embeddings to a dense neural network classifier (NN model). It has two layers, one with 256 hidden units and ``relu'' activation, and an output layer with the same number of units as classes in the data set, and ``softmax'' activation. We train the model using the adam optimizer, and categorical cross-entropy as the loss function. We hold out 10\% of the training data as a validation set. The model is trained for up to 100 epochs, with early stopping in case the model does not improve its validation accuracy for 30 epochs. Finally, we select the model as trained after the epoch with the highest accuracy on the validation split for use in reporting the results.

While we are most interested in how data set changes affect the  model's behavior, we note that the NN model seems to perform well. Given its rapid execution, it apparently benefits from the pre-trained sentence encoder and additional layer on top.



\subsection{Experiments}
\label{sec:experiments}
\label{sec:exp-e1}
{\bf E1 - Sample-Size Sensitivity:} In this experiment we analyzed the impact of increasing the size of a data set. For the BVA and ISC data sets, we performed the experiment at the document (i.e., a court opinion) level. Each data set has 50 documents. Since the StatInt data set features only one or a few sentences per case, we grouped them together in ``documents'' of 10 cases each, yielding 37 (pseudo) documents. We split all of the documents into a training and a test split, with a ratio of 80/20\%. In the first iteration, only the sentences coming from a single document in the training split were used to train the model. Then, for each iteration the sentences from one additional document were added. At each iteration, the model performance was evaluated against the test split on both an overall and a per-label basis. This allowed us to investigate how adding data to the training set impacts the performance of the classifier, and whether performance trends suggest that adding additional data could be beneficial.

{\bf E2 - Split Sensitivity:} It is common practice to divide the data set into training, validation, and test splits. However, it has been shown (e.g., \cite{Fuhr2018}) that the so-called holdout method of separating the available data into a training set and a (disjoint) test set can yield unreliable outcomes, especially when using comparatively small data sets. This is because the results may depend heavily on the split (e.g. the study in \cite{Rao2015} shows that the performance on the two halves of the collection differed by 20\%, and, for other splits, system differences sometimes were significant). Therefore, cross-validation is recommended (see e.g. \cite[p. 152--6]{Witten2016}). This problem is aggravated when we deal with data sets of smaller sizes, where the distribution of the training data is likely to differ substantially between different splits.

To investigate the impact of split selection, we split each data set into five folds. In each iteration, four of the folds were used as training split, and the remaining fold as test split. This allowed us to observe the particular split's impact on the performance of the trained NN model, and whether the scores vary overall and on a per-label basis.

{\bf E3 - Error Sensitivity:} The high-quality of the human labelling is an important concern in ML. The inevitable errors in human labelling could have a negative impact on the training, since the model could detect an incorrect pattern or receive confusing signals. To investigate the impact of labelling errors on classifier performance, we started with the data as published (see Section \ref{sec:data}). In each subsequent iteration, we replaced an increased percentage of labels in the training split (see Experiment 1 above) with randomly chosen incorrect labels. At each iteration, we evaluated the performance of the model against the test split, on both an overall and a per-label basis. As the percentage of incorrect labels in the training set increases, the performance of the trained NN model is likely to decrease. This experiment revealed the rate of the performance degradation.

Human errors from mislabeling are unlikely to be random. To investigate the effects of consistent mislabeling, we ran a second phase of this experiment. Instead of replacing gold labels with another random label, we replaced labels consistently with another label. Essentially, the first label was always replaced with the last label, the second by the first etc. For example, whenever an ``Other'' type was mutated in the BVA data set, it was replaced with a ``Reasoning'' sentence label. This emulates a scenario where certain labels are more likely to be confused with certain other labels.

\section{Results}

\begin{figure}
    \includegraphics[width=.3\textwidth]{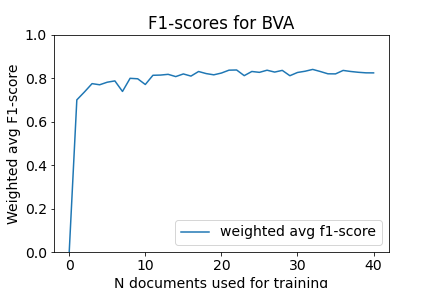}
    \includegraphics[width=.3\textwidth]{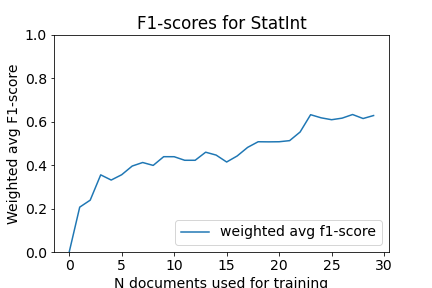}
    \includegraphics[width=.3\textwidth]{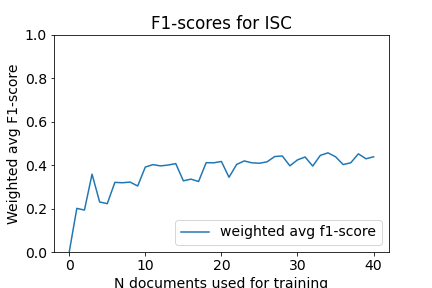}
    \caption{Evolution of overall WAVG F1-score of classifier, as documents added to training data one by one.}
    \label{fig:E1-overall}
\end{figure}

\begin{figure}
    \includegraphics[width=.3\textwidth]{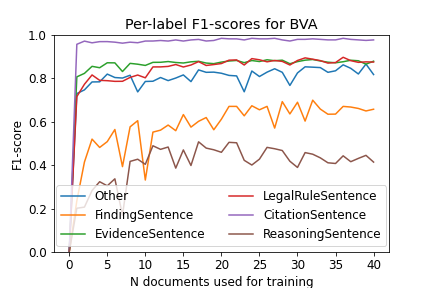}
    \includegraphics[width=.3\textwidth]{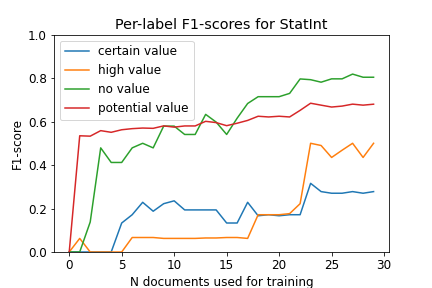}
    \includegraphics[width=.3\textwidth]{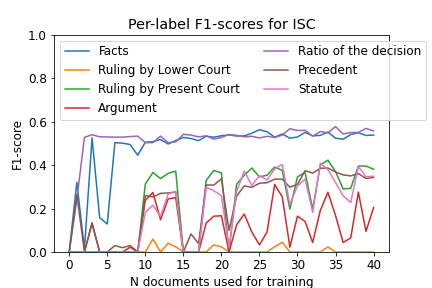}
    \caption{Evolution of per-label F1-score of classifier, as documents are added to  training data one by one.}
    \label{fig:E1-labels}
\end{figure}

{\bf E1 - Sample-Size Sensitivity:} Figure \ref{fig:E1-overall} shows the overall evolution of the weighted average F1-score when adding additional documents. The score from the BVA data set rises steeply almost immediately, reaches a fairly high level, and seems close to its maximum performance after incorporating all the documents. The scores for both StatInt and ISC rise more slowly, achieve lower levels, and show an upward trend in performance even when all the documents are included. This indicates that adding additional data to these data sets could further improve the performance.

Figure \ref{fig:E1-labels} shows the  evolution of F1-scores for each individual class. The rates of improvement among the  classes are not consistent. For the BVA data set, most of the labels reach a high level of performance quite quickly and then improve slowly. On the other hand, the ``FindingSentence'' and ``ReasoningSentence'' classes take more time to achieve their highest performance. For the StatInt data set, the ``potential value'' and ``no value'' classes rapidly achieve high performance, while the other two classes require more data to improve. For the ISC data set, the ``Ratio of the decision'' and ``Facts'' classes improve rapidly initially, while the other classes vary significantly in performance depending on the data added, while showing an overall positive trend. The amount of noise around the per-label trends also varies considerably, in all three data sets.

\begin{figure}
    \includegraphics[width=.3\textwidth]{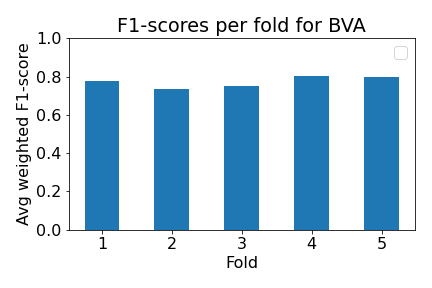}
    \includegraphics[width=.3\textwidth]{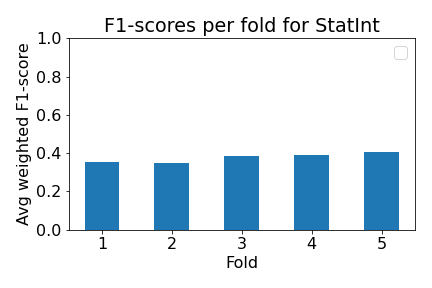}
    \includegraphics[width=.3\textwidth]{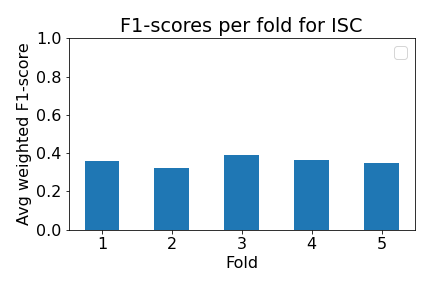}
    \caption{Variations in weighted average F1-score per five different folds for training and test data.}
    \label{fig:E2-overall}
\end{figure}

\begin{figure}
    \includegraphics[width=.3\textwidth]{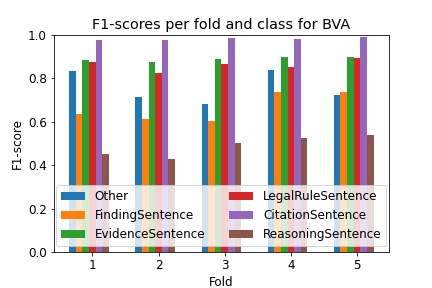}
    \includegraphics[width=.3\textwidth]{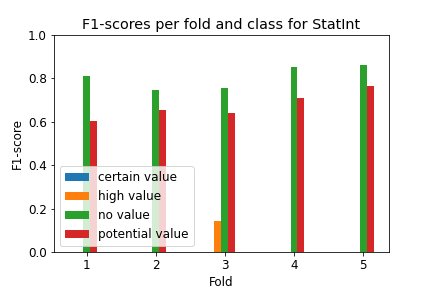}
    \includegraphics[width=.3\textwidth]{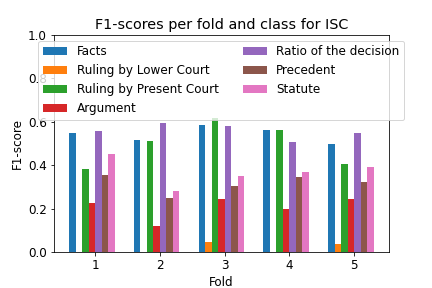}
    \caption{Variations in F1-score for individual classes per five different folds for training and test data.}
    \label{fig:E2-labels}
\end{figure}

{\bf E2 - Split Sensitivity:} Figure \ref{fig:E2-overall} shows the overall weighted average F1-score of the NN model trained on five different folds. We observe that the performance varies across the folds. Figure \ref{fig:E2-labels} shows the scores per class. The scores of some classes differ considerably across the folds. 
Some classes seem to perform rather consistently across the folds (e.g. ``Citation'', ``LegalRule'' and ``Evidence'' for BVA and ``no value'' for StatInt). Other classes show higher variation in performance (e.g. ``FindingSentence'' for BVA, ``high value'' for StatInt and ``Ruling by Present Court'' for ISC).

\begin{figure}
    \includegraphics[width=.3\textwidth]{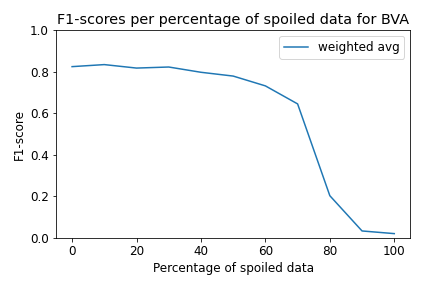}
    \includegraphics[width=.3\textwidth]{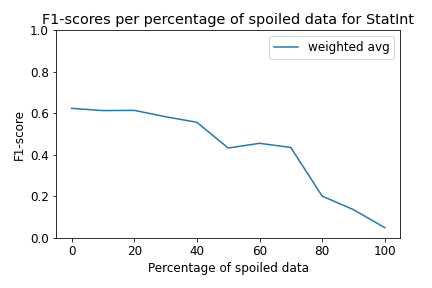}
    \includegraphics[width=.3\textwidth]{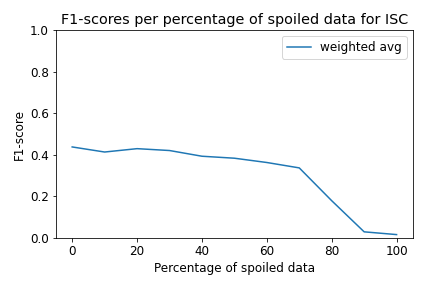}
    \caption{Evolution of weighted avg. F1-score when replacing  set percentage of labels with  random  label.}
    \label{fig:E3-random-overall}
\end{figure}

\begin{figure}
    \includegraphics[width=.3\textwidth]{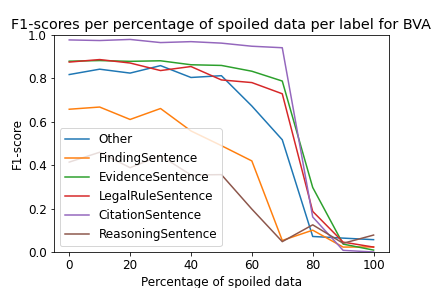}
    \includegraphics[width=.3\textwidth]{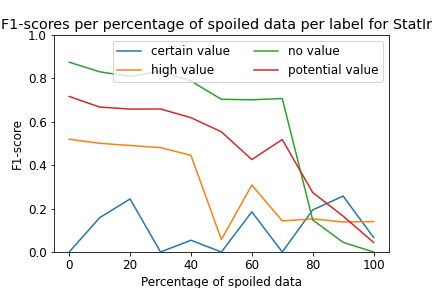}
    \includegraphics[width=.3\textwidth]{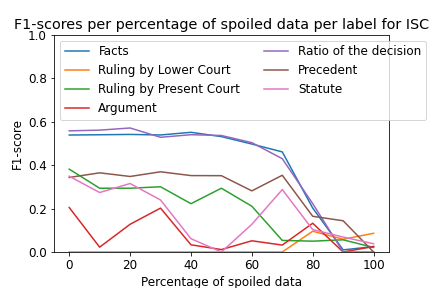}
    \caption{Evolution of per-class F1-scores when replacing set percentage of labels with random  label.}
    \label{fig:E3-random-labels}
\end{figure}

{\bf E3 - Error Sensitivity:} Figure \ref{fig:E3-random-overall} shows the effects of randomly mislabeling a portion of the training data set. While performance does begin to decrease immediately with the introduction of the erroneous labels, it remains surprisingly stable until around 70\% of the data has been replaced. Figure \ref{fig:E3-random-labels} shows the evolution of performance on a per-label basis. Some of the classes start to lose performance quite rapidly, while the others are more resilient to the random errors.

\begin{figure}
    \includegraphics[width=.3\textwidth]{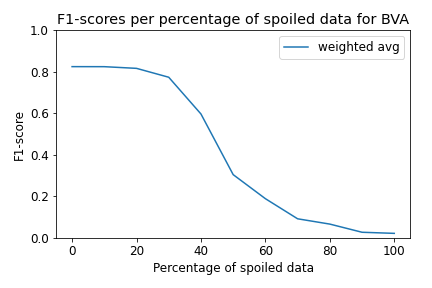}
    \includegraphics[width=.3\textwidth]{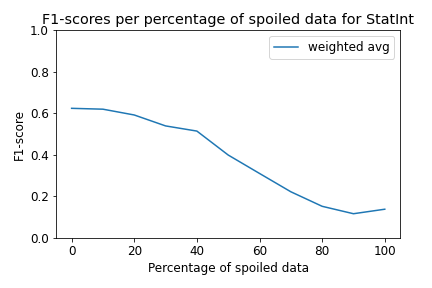}
    \includegraphics[width=.3\textwidth]{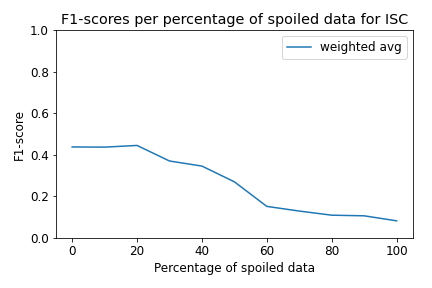}
    \caption{Evolution of WAVG F1-score when replacing set percentage of labels with consistent other label.}
    \label{fig:E3-consistent-overall}
\end{figure}

\begin{figure}
    \includegraphics[width=.3\textwidth]{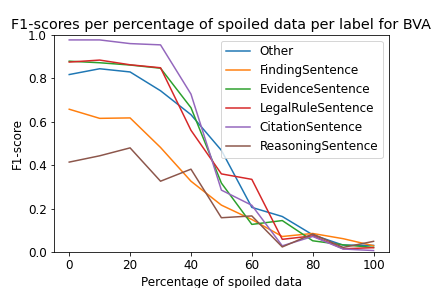}
    \includegraphics[width=.3\textwidth]{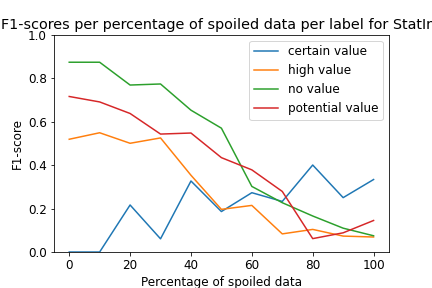}
    \includegraphics[width=.3\textwidth]{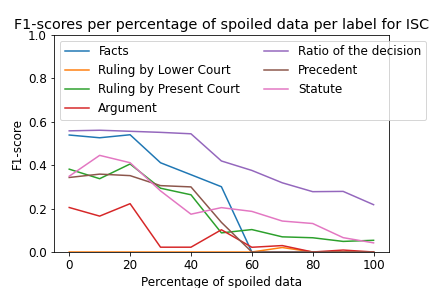}
    \caption{Evolution of per-class F1-scores when replacing  set percentage of labels with  consistent other label.}
    \label{fig:E3-consistent-labels}
\end{figure}

Figure \ref{fig:E3-consistent-overall} shows a different picture, as each label is replaced with the same label instead of a randomly chosen one. Here, the performance decreases much more quickly, with significant performance impacts being visible at around 25\% of the data being mislabeled. Comparing Figure \ref{fig:E3-consistent-labels} to Figure \ref{fig:E3-random-labels} shows that this pattern holds for the individual class scores, which begin decreasing much earlier than with the random replacement.

\section{Discussion}
In this paper, we altered the quantity, training/test split, and labelling quality of data sets to investigate the impacts on a trained NN classifier. The results yield a number of insights that could be useful for further research in the field of AI \& Law.


Overall, \textbf{Experiment 1} clearly shows the varying importance of larger data sets. For two of the data sets, adding additional data could further improve performance, while the third could achieve a fairly high level of performance using less data. Moreover, Figure \ref{fig:E1-labels} shows that in all three data sets the improvement is not uniform across the classes. While some of them immediately rise to a very high level of performance, others take  longer to do so. This suggests that, for any level of performance deemed adequate for a particular use case, the needed sample size can vary depending upon the class.


\textbf{Experiment 2} demonstrates how the overall performance and per-label performance differ across folds for training and test data. Both the overall f1-scores and some of the per-label scores show variations between the different folds. 
Depending on the use case, selecting a single train-test split may not produce reliable results when working with relatively limited legal data sets. Both the overall and per-label results may vary sufficiently that reporting results from a single fold could hide the true score of a model.


\textbf{Experiment 3} investigates the effects of mislabeled data on classifier performance. Surprisingly, the labels mutated in a random fashion in Figure \ref{fig:E3-random-overall} and Figure \ref{fig:E3-random-labels} do not seem to affect the performance of the classifier significantly until a large portion of the labels have been altered. Figure \ref{fig:E3-consistent-overall} and Figure \ref{fig:E3-consistent-labels} show that performance decreased more rapidly when we replaced the labels with a consistent erroneous label. In real-world scenarios,  human annotators may more likely make systematic errors, such as regularly mislabelling a ``Reasoning'' sentence as a ``Finding'' sentence in the BVA data set.

\textbf{Choice of metrics:} Across the experiments, we see a possible moment of confusion arising from the metrics used. While the weighted average F1-score fluctuates, the per-label statistics show that the fluctuation is much greater for some labels than others. This is likely due to the weighted average F1-score assigning a larger weight to classes with more samples, which are also less likely to be affected by the selection of training data. 

\textbf{Class difficulty and semantic homogeneity:} Taking the three experiments together,  with regards to performance on a per-label basis, it appears that certain labels are significantly more difficult for the classifier to learn than others. In Figure \ref{fig:E1-labels}, some labels (such as ``Citation'' for BVA, ``potential value'' for StatInt and ``Ratio'' for ISC) quickly achieve a high level of performance and then improve more slowly. Other labels require more data to achieve a higher performance, and improve throughout the experiment (such as ``Finding'' for BVA, ``high value'' for StatInt and ``Precedent'' for ISC). In Figure \ref{fig:E2-labels} the latter labels vary to a greater extent between different folds. Their performance also deteriorates more quickly when introducing erroneous labels, as in Figures \ref{fig:E3-random-labels} and \ref{fig:E3-consistent-labels}. Thus, it appears that some label classes are more difficult for the classifier to learn than others; they are more sensitive to data sparsity, training/test split and labelling errors.

A possible explanation for such divergence might be the \textit{frequency of the labels} in the data set. The classifier will have more opportunities to learn classes that appear more frequently. This certainly seems to explain some of the divergence. In the BVA data set, for example, the ``Evidence'' and ``Citation'' classes are the most common, and also achieve high scores very early in the process. However, class frequency does not explain all of the difference. For example, ``Other'' and ``LegalRule'' perform better than ``Finding'' and ``Reasoning'' in the BVA data set, despite containing fewer examples. 

Another possible explanation is the \textit{semantic homogeneity} of a class, i.e., how semantically similar the sentences are within a particular class. Certain classes, such as citations or word-for-word referrals to statutes, are likely to be more similar to each other, i.e., arranged in homogeneous clusters when embedded into a semantic space. Other classes, such as free-form reasoning sentences, might cover a wider semantic space, overlapping with sentences from other classes. 

\begin{figure}
    \includegraphics[width=.32\textwidth]{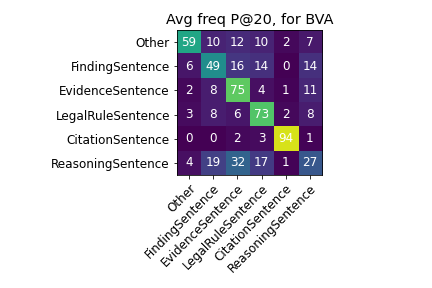}
    \includegraphics[width=.32\textwidth]{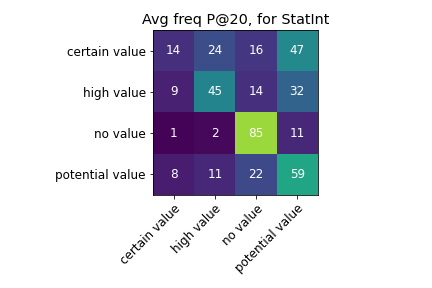}
    \includegraphics[width=.32\textwidth]{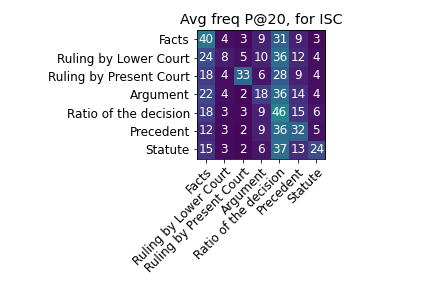}
    \caption{Frequency in percentages of labels in the top 20 most similar, per label. Rows represent the base label, while columns represent the average percentage of labels occurring in the top 20 most similar labels.}
    \label{fig:E4-labels}
\end{figure}

In \cite{westermann2020} an annotation tool was presented that grouped sentences based on semantic similarity in an embedding space (as determined by Euclidian distance in the embedding space). The goal was to enable more efficient annotation of legal documents. For each sentence in a certain class, we explored how many on average of the top 20 most similar sentences were also of that same class. A high number indicates a more semantically homogeneous class, while a lower number indicates that the class might be mixed with other classes. We employed the same methodology and metric here. Figure \ref{fig:E4-labels} presents the percentage of other labels that occur on average in the top 20 most similar sentences. For example, looking at Figure \ref{fig:E4-labels} and the ``FindingSentence'' row for BVA, we  see that on average 49\% of the 20 most semantically similar sentences were also of the class ``FindingSentence'', and the most confusion was with ``EvidenceSentence''. By contrast, the row for ``CitationSentence'' shows very high semantic homogeneity. 

When comparing the results from Figure \ref{fig:E4-labels} with the ``difficult'' classes in  our experiments, we see that the higher the semantic homogeneity of a class, the less difficult a class seems to be for the NN model to learn. For example, the BVA class ``CitationSentence'' has the highest prediction performance, it is largely unaffected by the chosen split, and it is relatively robust against labelling errors. The class ``ReasoningSentence'', with a semantic homogeneity of only 27\%, has the lowest final prediction performance in experiment 1, one of the higher sensitivities to the selected split, and begins losing performance relatively early when errors occur. This link between semantic homogeneity and class difficulty seems to hold up over most of the classes across the data sets.

Semantic homogeneity may thus be a significant factor why some classes (and data sets) perform better than others. Although determining  why this is the case is beyond the scope of this work, it seems likely that the classifier can more easily find decision boundaries for sentences grouped into clear semantic clusters.

\textbf{Tentative implications for AI \& Law research:} The experiments yield several insights that could be useful in AI \& Law. By adding data to a classifier in an iterative fashion, as we did in Experiment 1, we can understand the overall trend of how the classifier performance develops. This might support the decision about collecting additional data. Further, by analyzing the semantic homogeneity and class difficulty observed in the experiments, we can choose to deliberately focus data collection on specific classes, which can potentially improve overall performance. These insights show how focusing on the data set, rather than the model, can support researchers in understanding results and making choices that have the potential to significantly advance performance on a particular task.


\section{Conclusions and Future Work}
We trained a classifier on three data sets, altering the size, training/test split and data labelling quality, to investigate the effects of these properties on ML classifier performance. We observe significant variations in performance over the experiments. We identify potential reasons for these variations, and discuss how the experiments might be helpful for future research carried out in AI \& Law.



In future work, we plan to expand the experiments to additional data sets. The experiments might also be modified, such as performing the sample-size experiment on a sentence-level rather than a document-level, or introducing systematic errors based on the semantic confusion between classes (similar to Figure \ref{fig:E4-labels}). 
Finally, we have identified a need for a methodology on empirical investigation of the properties of data sets. Such an established methodology could be used to guide data collection and curation efforts and improve the quality of the data sets.
Potentially, the methodology could be used to compare the properties of different data sets with each other. While more research is needed on this front, our research opens a promising door towards such a methodology.


\begin{thebibliography}{99}


\bibitem{bhattacharya2019}
Bhattacharya, P., Paul, S., Ghosh, K., Ghosh, S. \& Wyner, A. ``Identification of Rhetorical Roles of Sentences in Indian Legal Judgments.'' \emph{Jurix 2019}. pp. 3-12 (2019).

\bibitem{cer2018}
Cer, D., Y. Yang, S. Kong, N. Hua, N. Limtiaco, R. St John, N. Constant et al. ``Universal sentence encoder.'' \emph{arXiv preprint arXiv}:1803.11175 (2018).

\bibitem{Fuhr2018}
Fuhr N. ``Some common mistakes in ir evaluation, and how they can be avoided.'' In \emph{ACM SIGIR Forum}, vol 51, pp. 32-41. ACM, (2018).

\bibitem{iyyer2015}
Iyyer, M., Manjunatha, V., Boyd-Graber, J., and Daum\'e III, H. ``Deep unordered composition rivals syntactic methods for text classification.'' In \emph{Proc. 53rd Ann. Mtg.  Assoc. for Computational Linguistics and 7th Int'l Joint Conf. on Natural Language Processing}, vol 1, 1681-1691 (2015).

\bibitem{nithya2021}
Sambasivan, N. et al. ``Everyone wants to do the model work, not the data work’’: Data Cascades in High-Stakes AI. \emph{2021 CHI Conference} pp. 1-15 (2021).


\bibitem{northcutt2020}
Northcutt, C. G., Athalye, A. \& Mueller, J. ``Pervasive Label Errors in Test Sets Destabilize Machine Learning Benchmarks.'' \emph{arXiv:2103.14749} [cs, stat] (2021).


\bibitem{Rao2015}
Rao, J., Lin, J. \& Efron, M.  ``Reproducible experiments on lexical and temporal feedback for tweet search.'' In \emph{European Conference on Information Retrieval}, pp. 755–767. Springer, (2015).

\bibitem{savelka2021}
Savelka, J.,  Westermann, H., Benyekhlef, K.  et al.
``Lex Rosetta: transfer of predictive models across languages, jurisdictions, and legal domains.'' In 
\emph{ICAIL 2021}, pp. 129–138. (2021). 

\bibitem{savelka2020}
Savelka, J., Westermann, H. \& Benyekhlef, K. ``Cross-Domain Generalization and Knowledge Transfer in Transformers Trained on Legal Data.'' \emph{ASAIL@ Jurix.} (2020).

\bibitem{savelka2019}
\v{S}avelka, J., Xu, H., \& Ashley, K. ``Improving Sentence Retrieval from Case Law for Statutory Interpretation.'' \emph{Proc. 17th Int'l Conf. on Artificial Intelligence and Law}, pp. 113-122. (2019).

\bibitem{sureshguttag2020}
Suresh, H., Guttag, J. V. ``A Framework for Understanding Unintended Consequences of Machine Learning''. \emph{arXiv:1901.10002 [cs, stat]}, (2020), \url{http://arxiv.org/abs/1901.10002}

\bibitem{tan2020}
Tan, H., Zhang, B., Zhang, H., \& Li, R. ``The Sentencing-Element-Aware Model for Explainable Term-of-Penalty Prediction''. In \emph{CCF Int'l Conf. on NLP and Chinese Computing}. pp. 16-27. (2020).  

\bibitem{vaswani:2017}
Vaswani, A., Shazeer, N., Parmar, N., Uszkoreit, J., Jones, L., Gomez, A.N., Kaiser, Ł. and Polosukhin, I., ``Attention is all you need.'' In \emph{Advances in neural information processing systems.} (2017).

\bibitem{walker:2019}
Walker, V. R., et al. ``Automatic Classification of Rhetorical Roles for Sentences: Comparing Rule-Based Scripts with Machine Learning.'' \textit{Proceedings of ASAIL 2019} (2019).

\bibitem{walker:2017}
Walker, V. R., et al. ``Semantic types for computational legal reasoning: propositional connectives and sentence roles in the veterans' claims dataset.'' \textit{Proceedings of ICAIL '17}. ACM, (2017).

\bibitem{westermann2021}
Westermann, H., Savelka, J., Walker, V., Ashley, K. \& Benyekhlef, K. ``Data-Centric Machine Learning: Improving Model Performance and Understanding Through Dataset Analysis.'' \emph{Jurix 2021}.

\bibitem{westermann2020}
Westermann, H., Savelka, J., Walker, V., Ashley, K. \& Benyekhlef, K. ``Sentence Embeddings and High-Speed Similarity Search for Fast Computer Assisted Annotation of Legal Documents.'' \emph{Jurix 2020}.

\bibitem{Witten2016}
Witten, I.H., Frank E., Hall, M.A. \& Pal, C.J. \emph{Data Mining:  Practical machine learning tools and techniques}. Morgan Kaufmann, (2016).
 
\bibitem{zhong2019}
Zhong, L., Zhong, Z., Zhao, Z., Wang, S., Ashley, K. D., \& Grabmair, M. ``Automatic summarization of legal decisions using iterative masking of predictive sentences.'' 
\emph{ICAIL 2019}, pp. 163-172. (2019).

\bibitem{zhou2020}
Zhou, X., Zhang, Y., Liu, X., Sun, C., \& Si, L. ``Legal Intelligence for E-commerce: Multi-task Learning by Leveraging Multiview Dispute Representation.'' In \emph{Proc. 42nd Int'l ACM SIGIR Conf. on Research and Development in Information Retrieval}, pp. 315-324. (2019).




\end{thebibliography}
\end{document}